\documentclass[
]{ceurart}

\usepackage{listings}
\usepackage{comment}

\usepackage{tikz}
\usetikzlibrary{calc}
\usetikzlibrary{arrows.meta, positioning}
\usepackage{wrapfig}
\newcommand{\Prob}{\mathrm{P}}

\begin{document}

\copyrightyear{2026}
\copyrightclause{Copyright for this paper by its authors.
  Use permitted under Creative Commons License Attribution 4.0
  International (CC BY 4.0).}

\conference{13th Italian Workshop on Artificial Intelligence and Robotics (AIRO 2026)}

\title{Estimating the Causal Impact of Social Navigation Signals in Corridor Scenarios}

\author[1]{Francesco Baldo}[%
orcid=0009-0009-9136-3619,
email=francesco.baldo.5@studenti.unipd.it
]
\author[1]{Luca Castri}[%
orcid=0000-0003-2525-8190,
email=lucacastri94@gmail.com
]

\author[2,1]{Gloria Beraldo}[%
orcid=0000-0001-8937-9739,
email=gloria.beraldo@{cnr, unipd}.it
]

\author[1]{Nicola Bellotto}[%
orcid=0000-0001-7950-9608,
email=nicola.bellotto@unipd.it
]

\address[1]{Department of Information Engineering, University of Padua}
\address[2]{Institute of Cognitive Sciences and Technologies, National Research Council of Italy}

\begin{abstract}
Deploying robots in human-shared environments requires deciding not only how to navigate, but also when dedicated social signals are worth emitting. Classic observation-based inference can be misleading in such settings: confounding factors may bias the estimated relationship between a signal and the task outcome, leading to incorrect decisions.
Causal inference provides a framework for disentangling these relationships by modelling the mechanisms linking observations, actions, and outcomes, and by estimating interventional effects after accounting for confounding paths.
We study a human-robot interaction scenario in a corridor and use a causal model to estimate whether a social navigation signal causally improves transit success. We represent the interaction as a causal DAG involving robot action, pedestrian occupancy, available space, task outcome, and a controlled confounder (static obstacles). We compare the interventional estimate with the observational estimate obtained by conditioning only on the robot action, which does not explicitly model interventions.
Our results show that accounting for confounding variables can reverse the conclusions drawn from naïve observational analysis: the observational estimate wrongly suggests that the social signal is detrimental, whereas the causal estimate shows that it is beneficial. This discrepancy highlights how failing to account for confounding can lead to incorrect conclusions and demonstrates the importance of causal models for context-aware signalling decisions in social robot navigation.
\end{abstract}

\begin{keywords}
  Causal reasoning \sep
  Social navigation \sep
  Human-Spatial Interaction\sep
  Mobile robotics
\end{keywords}

\maketitle

\section{Introduction}
Successful human-robot navigation in shared spaces depends not only on collision avoidance, but also on the ability of both agents to understand each other's intentions \cite{bacchin2025preference, gebelli2025dynamics, love2026hexar}. While humans naturally communicate their future actions through subtle changes in movement, robots cannot rely on the same implicit communication mechanisms. This asymmetry makes intention communication a fundamental requirement for achieving fluid and socially acceptable interactions.

Consider two people meeting in a narrow corridor. In many cases they resolve the encounter within about a second and without exchanging a word: one slightly reduces speed, the other reads this as intent and steps aside. Neither is merely an obstacle to the other, but an active participant whose motion is inherently expressive, with changes in speed and direction conveying where a person is going and what they are about to do \cite{kendon1977studies}.

\begin{figure}[t]
\centering
  \includegraphics[width=0.6\textwidth]{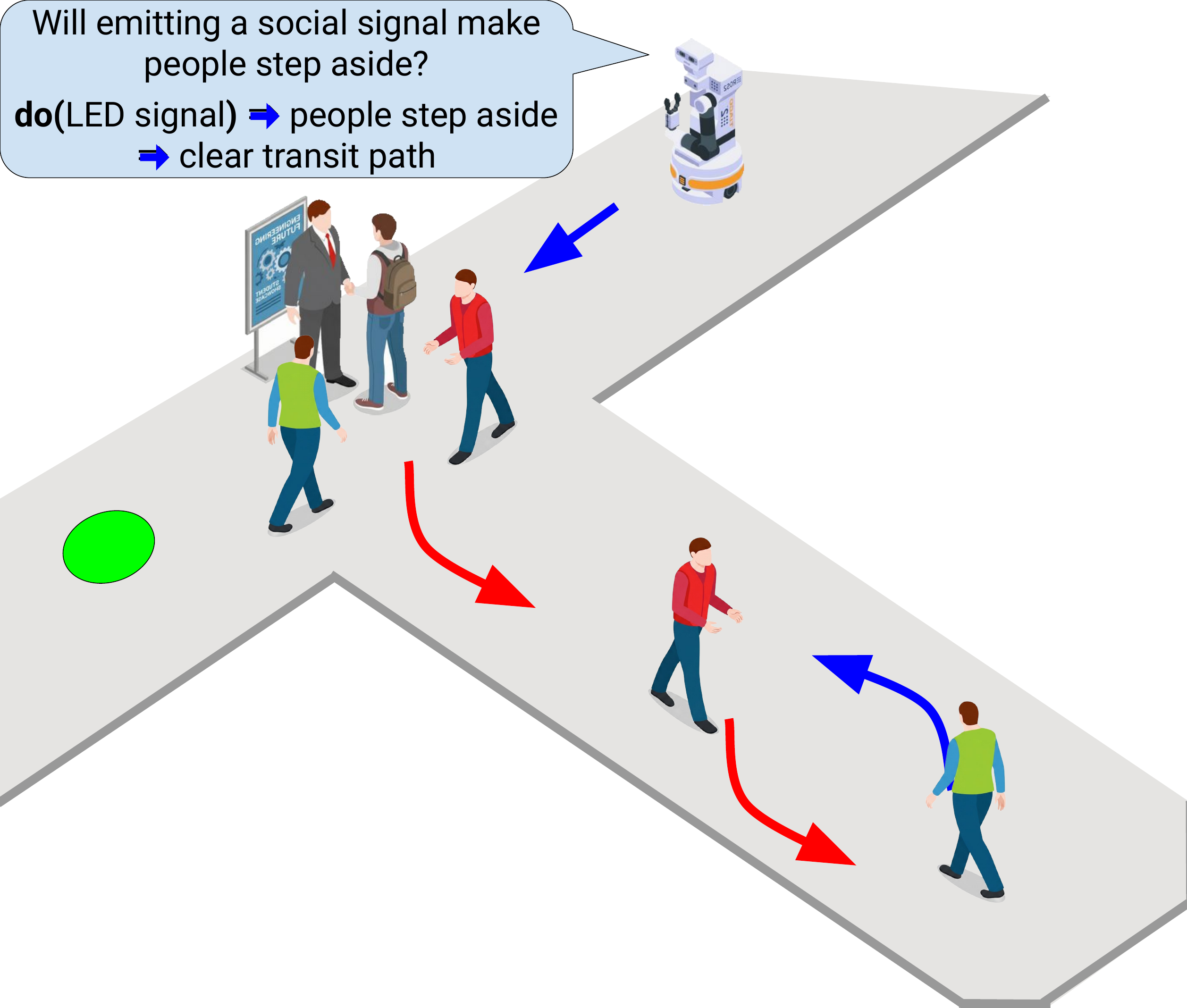}
    \caption{Motivating scenario: while humans naturally resolve encounters through subtle implicit cues, a robot must evaluate the causal impact of its intent signalling (e.g., LEDs blink) to determine whether emitting a signal genuinely prompts pedestrians to step aside and clear its path to the goal (green dot).}
\label{fig:intro}
\end{figure}

A social robot in the same corridor, as shown in Figure~\ref{fig:intro}, cannot rely on these mechanisms~\cite{mumm2011human}. Its navigation system typically models people as dynamic obstacles to be predicted and avoided, while its own planned motion does not clearly communicate its intentions~\cite{dragan2013}, making coordination less intuitive than in human-human encounters. Many approaches therefore introduce dedicated communication channels, such as directional light signals conveying navigation intent~\cite{szafir2015,mavrogiannis2023}. However, their effectiveness depends not only on communicating intentions, but on the robot's ability to determine when this communication is actually relevant.

In many situations, the robot must distinguish between factors that are genuinely relevant for decision-making and variables that only appear associated with the desired outcome. Confounding factors can lead to misleading associations, for example by suggesting that the presence of a person always implies the need for an explicit interaction signal, even when the person is not affected by the robot's behaviour. Causal reasoning ~\cite{pearl2009} provides a framework to disentangle these relationships by modelling the mechanisms that link observations, actions, and outcomes, enabling more informed and context-dependent decisions.

Building on this motivation, this paper asks not whether a signal is generally associated with successful interactions, but whether emitting it causally influences the outcome once contextual confounders are taken into account. We model the interaction between robot, environment and pedestrians as a causal graph and estimate the effect of signalling on observational data, as a proof-of-concept of how confounding can reverse the conclusions of a naïve analysis.

\section{Related work}

Although causal reasoning has become an increasingly active research area in robotics, its adoption remains uneven across different domains. Hellström~\cite{hellstrom2021} provides a broad overview of the field, organised around a categorisation of robot causal cognition mapped onto the sense-plan-act level of robot architectures, and observes that most of the resulting categories are supported by relatively little published work: causal notions are often implicit in the way robotic problems are formulated, but are rarely stated or exploited as such. 
Robotics is nonetheless a favourable setting, since a robot acts on its environment and can be instructed to act differently: interventional data arise from ordinary operation, whereas elsewhere they are expensive or unavailable.

To date, most work has focused on manipulation. Lee et al.~\cite{lee2021}
perform interventions within an approximate internal simulator in order to
establish which state variables actually affect the outcome of a task, and train
policies on the reduced state space alone. The resulting policies are
more sample efficient as well as more robust to domain shift and better able to
scale to settings in which many candidate variables are available. Planning
under uncertainty raises a complementary problem. Cannizzaro and
Kunze~\cite{cannizzaro2023} observe that POMDP planners lack explicit causal
semantics, so that in the presence of latent confounders the values they compute
are biased and the resulting policies underperform; they extend an online
planner with a causal model of the domain, whose parameterisation is partly
learned offline, and obtain policies that are robust to this bias. Causal models have also been applied to failure handling, where a causal Bayesian network learned in simulation and transferred to the real world allows a robot to explain why an action failed, predict whether it is likely to succeed, and select a corrective action to prevent the failure~\cite{diehl2023}.

In the context of human-robot interaction, the question becomes whether the robot influences the people around it at all. Tung et al.~\cite{tung2024} propose an information-theoretic approach to identify the states in which a robot's behaviour influences nearby humans, addressing the preliminary question of when such influence exists; this measure adapts to the multi-agent setting a criterion originally introduced for single-agent reinforcement learning~\cite{seitzer2021}. 
Evaluated on pedestrians simulated with the Social Force Model in a hallway, the
approach recovers the states in which the social forces of two agents overlap. Our work takes the existence of such influence as given in order to ask a further question: by how much a specific robot action changes the outcome of the interaction.

Recent studies have investigated the use of causal discovery to model and predict human spatial behaviour~\cite{castri2022}, improve scalability through feature selection in dynamic environments~\cite{castri2023}, and integrate the resulting causal models into an end-to-end framework for causal discovery and reasoning in robot navigation within human-shared environments~\cite{castri2026causality}. Moreover,~\citet{castri2024roscausal} introduced \textit{ROS-Causal}, a ROS-based causal analysis framework that enables data collection and causal discovery onboard the robot using its own sensor data.

While these works employ causal models as a component of the navigation system
to improve navigation performance, we instead use them to assess the value of a
specific robot action, comparing na\"ive and causally adjusted estimates of the
effect of a social navigation signal on task success over the same dataset.

\section{Causal Impact Estimation of Social Navigation Signals}
\label{sec:approach}

We first describe the human–robot spatial scenario used for our study, then formalize the interaction within a causal inference framework.

\subsection{Human–Robot Spatial Interaction in a Corridor Scenario}\label{sec:model}

To study the effectiveness of social navigation signals, we consider an indoor human-robot encounter at a spatial bottleneck. The goal is to evaluate whether emitting an explicit intent signal causally improves the robot's navigation success when traversing congested areas.

\begin{figure}[!ht]
  \centering
  \includegraphics[width=0.4\textwidth]{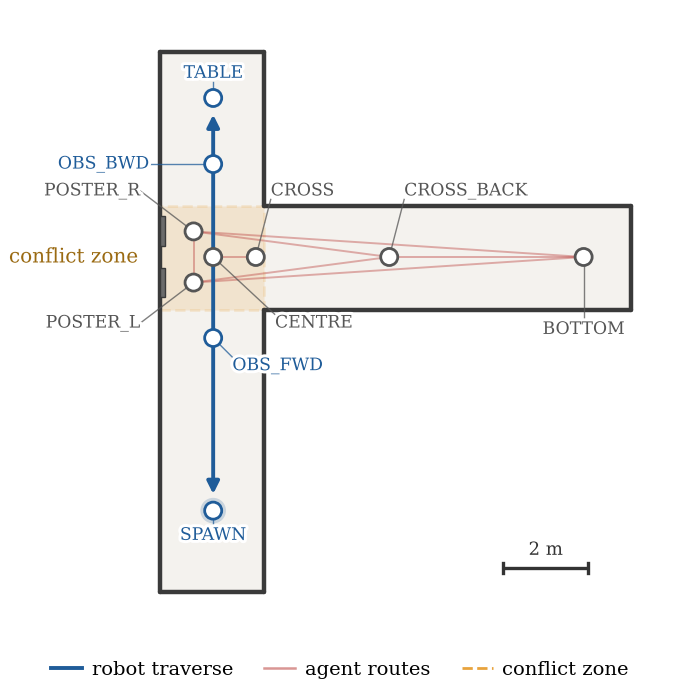}\hfill%
  \includegraphics[width=0.4\textwidth]{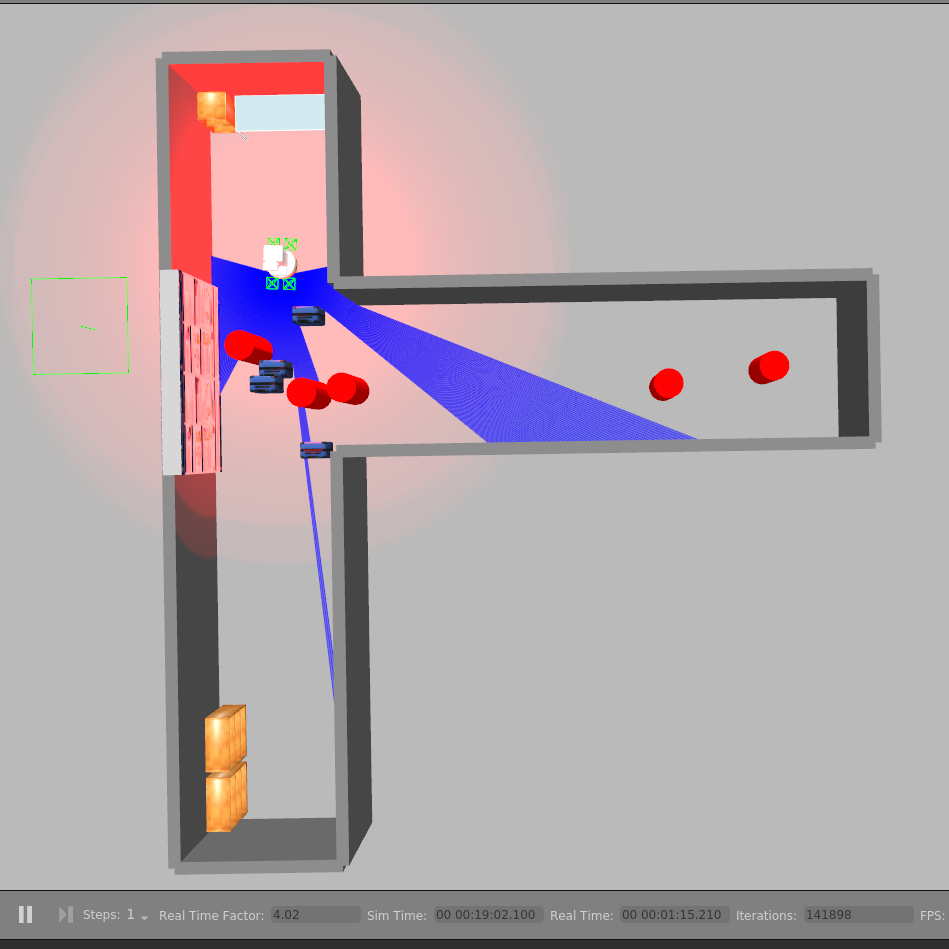}
  \caption{The corridor scenario. Left: layout and waypoint graph, with the robot path from \textsc{spawn} to
    \textsc{table}, the observation waypoints where the decision is taken, and the conflict zone at the junction
    (shaded). Right: the same corridor rendered in Gazebo, with pedestrians drawn as red cylinders and the static
    obstacles that constitute the confounder $O$ for the current episode.}
  \label{fig:scenario}
\end{figure}

The scenario, shown in Figure~\ref{fig:scenario}, consists of a T-shaped corridor, representative of environments such as universities, hospitals, and shopping malls. Both branches are $2.5$\,m wide: the main branch, along which the robot travels, is $12.8$\,m long, while the side branch opening onto it is $8.65$\,m long. The central junction, where the two branches meet through a $2.5$\,m-wide opening, constitutes the main bottleneck: the robot can traverse it only when pedestrians and other obstacles leave sufficient free space.

Five pedestrian agents populate the corridor with structured traffic routines designed to generate continuous spatial conflicts at the junction: two agents move within the conflict zone (shaded in Figure~\ref{fig:scenario}); a third agent repeatedly walks between $\mathrm{BOTTOM}$ and the junction, generating background traffic; and two agents travel between $\mathrm{BOTTOM}$ and the poster areas ($\mathrm{POSTER}_L$, $\mathrm{POSTER}_R$).

As the robot traverses the main branch between $\mathrm{SPAWN}$ and $\mathrm{TABLE}$, it must navigate through this central junction where pedestrian trajectories frequently cross, creating potential conflicts. Specifically, on each traversal the robot stops at the observation waypoint for that direction ($\mathrm{OBS\_FWD}$ or $\mathrm{OBS\_BWD}$). Here, it evaluates the passage congestion and chooses between two candidate actions: emitting a directional light signal requesting nearby pedestrians to make way ($A=1$), or proceeding to cross through $\mathrm{CENTRE}$ without signalling ($A=0$).

\subsection{Causal Impact Estimation}\label{sec:causal-impact}

To estimate the true interventional effect of emitting a social navigation signal on overall task completion, we formalize the interaction as a causal Directed Acyclic Graph (DAG). The DAG is not learned from data: it is a generative model of the scenario, specified from the mechanisms implemented in simulation, which is then run to produce the dataset of Sec.~\ref{sec:experiments}.

Two kinds of entity occupy the passage. Pedestrians perceive the signal and can step aside, so the action can change the space they take up. Static objects (suitcases, bags, boxes left in a corridor) occupy the same space and are sensed in the same way, but no signal will move them. At decision time the robot measures a single gap and cannot tell them apart, yet the outcome depends on them differently: this is precisely the asymmetry the causal model must disentangle.

The interaction is modeled as a DAG over six binary variables (Figure~\ref{fig:dag}):

\begin{figure}[b]
\centering
\begin{tikzpicture}[
  node distance = 14mm and 16mm,
  var/.style = {circle, draw, minimum size=8mm, inner sep=1pt, font=\small},
  edge/.style = {-{Stealth[length=2mm]}, thick},
  conf/.style = {-{Stealth[length=2mm]}, thick, dashed}
]
  \node[var] (Pi) {$P_i$};
  \node[var, right=of Pi] (A)   {$A$};
  \node[var, right=of A]  (Pe) {$P_e$};
  \node[var, right=of Pe] (S)   {$S$};
  \node[var, right=of S]  (T)   {$T$};
  
  \node[var, below=0.6cm of Pe] (O) {$O$};

  \draw[edge] (Pi) -- (A);
  \draw[edge] (A)  -- (Pe);
  \draw[edge] (Pe) -- (S);
  \draw[edge] (S)  -- (T);

  \draw[conf] (O) -- (A);
  \draw[conf] (O) -- (S);
\end{tikzpicture}
\caption{Causal DAG of the corridor scenario. The main causal chain runs $P_i \rightarrow A \rightarrow P_e \rightarrow S \rightarrow T$. The controlled confounder $O$ (dashed) enters at two points, opening the backdoor path $A \leftarrow O \rightarrow S \rightarrow T$.}
\label{fig:dag}
\end{figure}
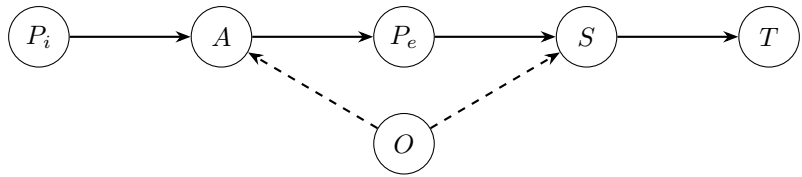

\begin{itemize}\setlength{\itemsep}{0pt}\setlength{\parskip}{0pt}
    \item \textbf{Confounder ($O$):} Presence of static, non-human obstacles such as luggage or bags, which occupy space as a pedestrian does but do not respond to the signal (1 = present, 0 = absent).
    \item \textbf{Initial Occupancy ($P_i$):} Crowd state before action ($1 = \text{congested}$, $0 = \text{open}$).
    \item \textbf{Robot Action ($A$):} Signal emission ($1 = \text{light ON}$, $0 = \text{light OFF}$).
    \item \textbf{Post-Action Occupancy ($P_e$):} Crowd state after pedestrian response time ($1 = \text{congested}$, $0 = \text{open}$).
    \item \textbf{Geometric Clearance ($S$):} Available transit gap (1 = sufficient, 0 = insufficient).
    \item \textbf{Task Outcome ($T$):} Final navigation result ($1 = \text{success}$, $0 = \text{timeout}$).
\end{itemize}

Each variable is recorded during an episode at the moment it becomes available, in the order of Figure~\ref{fig:timeline}: $O$ is drawn at the start of the traversal; at the observation waypoint the robot reads $P_i$ over pedestrians alone and selects $A$ on the gap left by pedestrians and obstacles together; $P_e$ re-evaluates $P_i$ once pedestrians have had time to respond; $S$ is computed immediately prior to transit over pedestrians, obstacles and walls; and $T$ records whether the robot reaches the goal within the time limit.

The main causal mechanism forms the chain $P_i \rightarrow A \rightarrow P_e \rightarrow S \rightarrow T$. Direct edges $A \rightarrow S$ and $P_i \rightarrow T$ are omitted because the action alters space solely by redirecting pedestrians, and initial crowd congestion influences outcome strictly through the robot's decision and subsequent crowd reaction. 

Static obstacles enter the graph at two distinct points. Through $O \rightarrow A$, obstacles reduce perceived space at decision time, increasing the likelihood of emitting a signal. Through $O \rightarrow S$, obstacles occupy physical space during transit, lowering success rates regardless of robot action. Conversely, the edge $O \rightarrow P_e$ is absent by construction: obstacles are invisible to the pedestrian simulator, ensuring that $P_e$ responds to the signal alone.

These two edges open a spurious backdoor path from action to outcome, $A \leftarrow O \rightarrow S \rightarrow T$, so evaluating $\Prob(T \mid A)$ without controlling for $O$ falsely associates signal emission with lower task success.
This is the only backdoor path from $A$ to $T$, and conditioning on $O$ blocks it without blocking any directed path from $A$ to $T$, so the set $\{O\}$ satisfies the backdoor criterion. Since we sample $O$ ourselves and record it in every episode, the criterion is not only satisfied, but also usable, and the causal effect is identified through backdoor adjustment:
\begin{equation} 
\label{eq:backdoor}
\Prob(T = 1 \mid \mathrm{do}(A = a)) = \sum_{o \in \{0,1\}}
\Prob(T = 1 \mid A = a, O = o)\, \Prob(O = o)
\end{equation}
This adjustment isolates the true interventional impact of the navigation signal by unconfounding the context in which the decision was taken.

\section{Experiments}\label{sec:experiments}

We simulate the corridor interaction across 100 traversal episodes and estimate the effect of signalling on transit success, comparing the naïve and adjusted estimators on the same data.

\subsection{Experimental Setup}
\begin{figure}[!ht]
    \centering
    \includegraphics[width=0.8\linewidth]{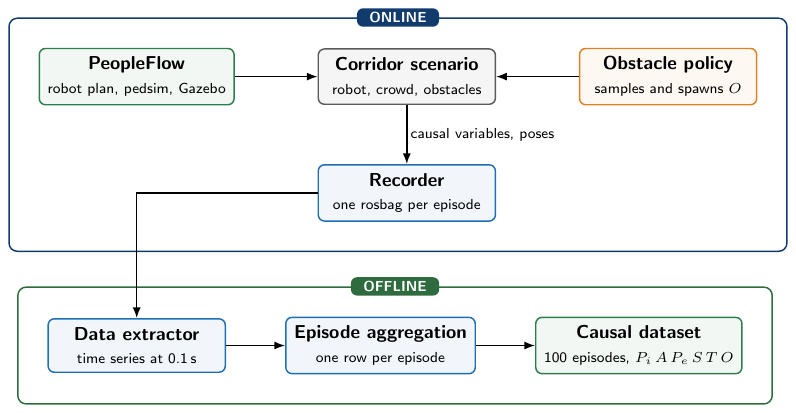}
    \caption{System architecture. Online (top): the robot plan drives the episode
       loop, coordinating the pedsim bridge, the obstacle policy and the recorder
       through ROS topics and parameters. Solid arrows indicate data flows, dashed arrows indicate
       parameters and control. Offline (bottom): the extraction pipeline turns the
       recorded bags into the per-episode causal dataset.}
    \label{fig:arch}
\end{figure}

Figure~\ref{fig:arch} shows the components underlying the system architecture, which is divided into two stages: (a) an online stage, in which the robot runs the episodes and records them, and (b) an offline stage, in which the recordings are processed to obtain the dataset used for estimation.

We implement a simulation world representing the corridor scenario introduced in Sec.~\ref{sec:model} (Figure~\ref{fig:scenario}), augmented with a mechanism for injecting a controlled confounder. Pedestrian dynamics follow the five-agent traffic routines defined in Sec.~\ref{sec:model} and are simulated in Gazebo via \textit{PeopleFlow}~\cite{castri2026causality}. Crowds are driven by \texttt{pedsim\_ros}\footnote{\url{https://github.com/srl-freiburg/pedsim_ros}}, which implements a Social Force Model coupled with a scheduling bridge that assigns agent destinations on request. 
The setup includes a simulated TIAGo robot\footnote{\url{https://pal-robotics.com/robot/tiago/}} equipped with a standard 2D navigation stack (global planner, TEB local planner, and AMCL localization). Robot behavior is orchestrated using PetriNetPlans\footnote{\url{https://github.com/iocchi/PetriNetPlans}}, which drives the episode loop and publishes each causal variable on a dedicated ROS topic exactly once per episode in causal order.

When the robot triggers the intervention ($A = 1$), it emits an LED signal and flags the pedestrians located within the look-ahead window, inside the conflict zone, or approaching the junction. The PeopleFlow bridge, which assigns each agent its next destination on request, returns for a flagged agent one of two predefined evasion waypoints instead of its scheduled destination, and the
Social Force Model walks the agent there. Compliance is deterministic; consequently, variation across episodes arises solely from which agents occupy the relevant areas at decision time. At the end of each episode, the flags are cleared and agents return to their default routes before the next traversal begins.

Two supporting nodes interface with the PeopleFlow simulator, synchronized through episode-boundary signals (Figure~\ref{fig:arch}): the obstacle policy, which samples $O$ at the beginning of each episode and, when $O = 1$, spawns four to seven suitcases at random positions within the junction, removing them at the end; and the recorder, which stores one ROS bag per episode, processed offline into a single observation tuple $(P_i, A, P_e, S, T, O)$. This representation is uniquely defined because each causal variable is published exactly once per episode in causal order.

\begin{figure}
    \centering
    \includegraphics[width=0.75\linewidth]{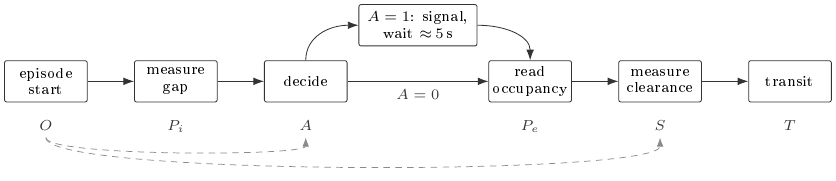}
    \caption{Timeline of an episode. Boxes are the phases of the protocol; below each, the causal variable recorded at that point. On $A=1$ the robot signals, redirects the affected agents and waits about five seconds before reading $P_e$; on $A=0$ the reading is immediate. The gap measured before the decision and the clearance measured before transit both include the static obstacles, whereas $P_i$ and $P_e$ are computed over pedestrians alone; the dashed arcs mark the two points where the confounder $O$ enters, as in Figure~\ref{fig:dag}.} 
    \label{fig:timeline}
\end{figure}

An episode corresponds to a single traversal, either from \textsc{spawn} to \textsc{table} or from \textsc{table} to \textsc{spawn}, alternating the two directions; Figure~\ref{fig:timeline} illustrates the episode timeline and indicates when each causal variable is recorded. When the passage is congested at decision time, the robot re-evaluates the gap before crossing. If the passage remains congested, it waits an additional 10 seconds before measuring the clearance again. The episode is considered successful if the robot reaches the goal, unsuccessful if any corridor traversal exceeds a fixed time threshold. In the event of failure, the robot returns to the starting waypoint and repeats the traversal in the same direction.

The robot always \emph{attempts} the transit, whatever the measured clearance
$S$: aborting when clearance is insufficient would impose the $S \to T$
relationship by design rather than let it be observed. Also, after signaling, it waits about five seconds before recording $P_e$ (Figure~\ref{fig:timeline}), so that redirected pedestrians have enough time to clear the path before the crowd state is reassessed.

We compute lateral clearance geometrically over a look-ahead window spanning $3.0$\,m in front and $0.5$\,m behind the robot across the full corridor width. Agents and obstacles are assigned a personal-space margin of $0.30$\,m in the projection. The perceptual threshold $G_\mathrm{soc} = 1.15$~m (robot width plus twice the personal margin) governs action selection: when the gap left by pedestrians and obstacles together falls below $G_\mathrm{soc}$, the passage counts as congested and the signal fires with probability 0.65; otherwise, the robot proceeds without signalling. The occupancy variables $P_i$ and $P_e$ are read with the same procedure and the same threshold, but over pedestrians alone, so that the crowd contribution to the gap is recorded separately from the obstacles. The physical threshold $G_\mathrm{phys} = 0.80$~m is stricter and binarizes the clearance $S$ available for the transit, where what matters is whether the robot has room to pass. 

A total of $100$ observational episodes were collected with no missing values or discarded runs. The marginal distributions are $\Prob(P_i{=}1) = 0.48$, $\Prob(A{=}1) = 0.38$, $\Prob(P_e{=}1) = 0.25$,
$\Prob(S{=}1) = 0.73$, $\Prob(T{=}1) = 0.76$ and $\Prob(O{=}1) = 0.29$.

The empirical confounder distribution matches the sampling probability, i.e. $\Prob(O = 1) \approx 0.30$. Furthermore, the interval between the two occupancy readings ($P_i$ and $P_e$) is below $0.1$\,s when untreated ($A = 0$), where there is no crowd response to wait for, and $5.6$\,s when treated ($A = 1$), confirming that the protocol provides the response delay only upon signalling.
An episode is a single traversal, outward from \textsc{spawn} to \textsc{table} or back, and the robot alternates the two directions. Action and success rates are virtually identical on the two legs, with $\Prob(A = 1) \in [0.37, 0.39]$ and $\Prob(T = 1) \in [0.75, 0.78]$, so we analyse them as a single dataset.

\subsection{Evaluation Metric \& Comparison Baseline}\label{sec:metrics}

To evaluate the causal efficacy of the social navigation signal, our primary metric is the Average Treatment Effect ($\mathrm{ATE}$) on task success $T$. The $\mathrm{ATE}$ measures the net change in task completion probability directly attributable to emitting the social signal, isolating the robot's intervention from contextual confounding. Specifically, we compared two impact estimation approaches:
\begin{itemize}
    \item \textbf{Na\"{\i}ve Observational Baseline:} 
    \begin{equation}\label{eq:bayes-ate}
    \mathrm{ATE}_{\text{naïve}} = \Prob(T = 1 \mid A = 1) - \Prob(T = 1 \mid A = 0)
    \end{equation}
    This computes the difference in conditional probabilities without adjustment and reflects standard observation-based inference, which does not model interventions and conflates correlation with causation in the presence of confounders.
    \item \textbf{Causal Interventional Approach:}
    \begin{equation}\label{eq:causal-ate}
    \mathrm{ATE}_{\mathrm{causal}} = \Prob(T = 1 \mid \mathrm{do}(A = 1)) - \Prob(T = 1 \mid \mathrm{do}(A = 0))
    \end{equation}
    This estimates $\mathrm{ATE}$ by computing the interventional distributions via backdoor adjustment on the confounder $O$ (Eq.~\ref{eq:backdoor}), effectively blocking the spurious backdoor path $A \leftarrow O \to S \to T$.
\end{itemize}
Both na\"{\i}ve and causal metrics were computed using pyAgrum\footnote{pyAgrum~\url{https://pyagrum.readthedocs.io/en/3.0.0/}}.

\section{Results}\label{sec:res}

\subsection{Na\"{\i}ve vs. Causal Interventional Estimation}\label{sec:naive_vs_causal}
\begin{figure}
    \centering
    \includegraphics[width=1\linewidth]{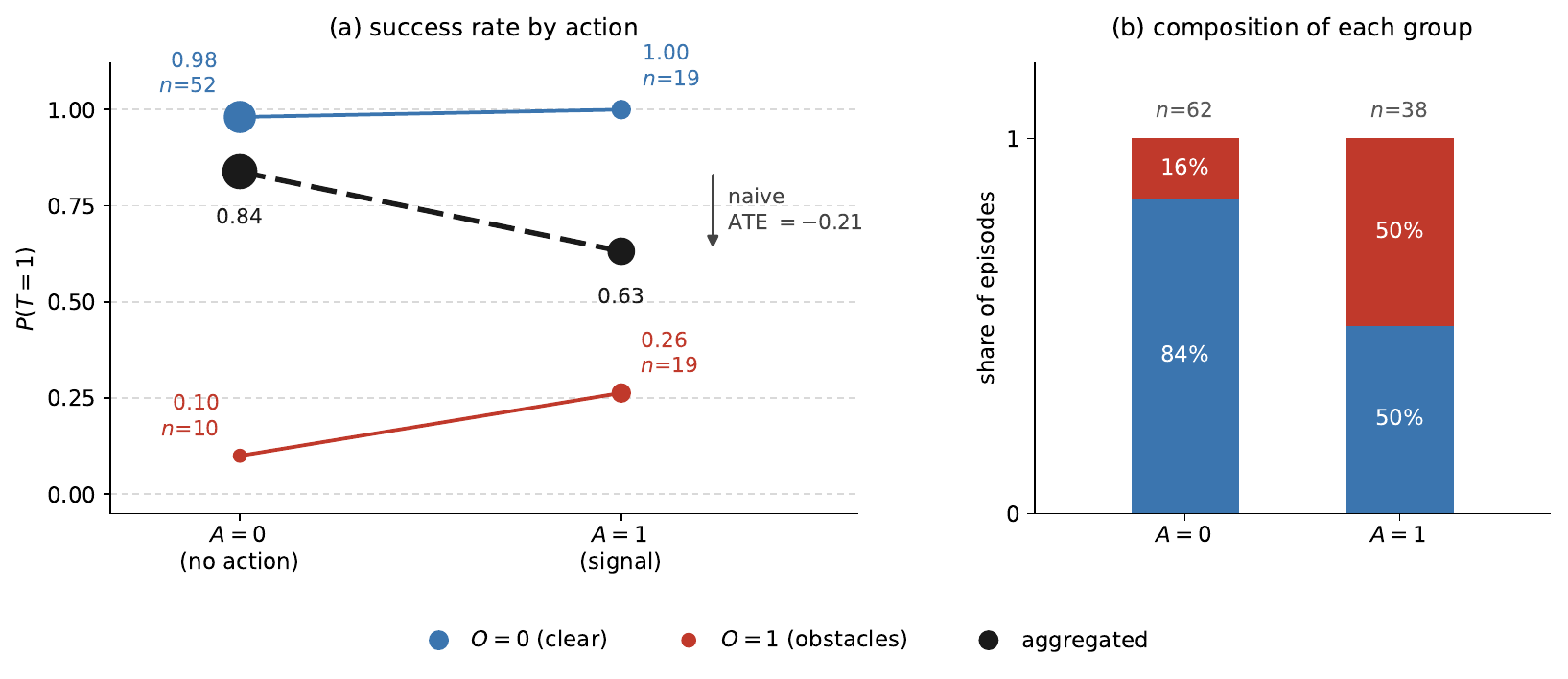}
    \caption{Comparison between na\"{\i}ve and causal estimate. (a) Success rate by action: within each level of $O$ the signal does not lower success, and in the obstructed context it raises it from $0.10$ to $0.26$, while the aggregate (black, dashed) falls by 21 points. (b) Composition of the two action groups: obstructed episodes are 16\% of those without a signal and 50\% of those with one, and that imbalance is what inverts the aggregate.}
    \label{fig:simpson}
\end{figure}
We first compare the na\"{\i}ve observational estimate of signal efficacy against the causally adjusted interventional effect across the $100$ recorded episodes. 

Evaluating task success conditionally on the robot's action alone yields:
\begin{equation}
\Prob(T = 1 \mid A = 1) = 0.63, \quad \Prob(T = 1 \mid A = 0) = 0.84 \implies \mathrm{ATE}_{\mathrm{naïve}} = -0.21
\end{equation}
Taken at face value, this na\"{\i}ve calculation suggests that activating the social signal is actively detrimental, reducing the navigation success rate by over $20$ percentage points. However, this association is spurious and entirely driven by environmental confounding.

Applying backdoor adjustment via Eq.~(\ref{eq:backdoor}) to the conditional frequencies $\mathrm{P}(T \mid A, O)$ weights the two contexts by the marginal distribution of $O$, giving $\mathrm{ATE}_{\mathrm{causal}} = 0.71 \times 0.02 + 0.29 \times 0.16 = +0.06$. The sign reverses: once the confounding effect of static obstacles is controlled for, emitting the signal is beneficial rather than harmful. This reversal reflects a classic manifestation of Simpson's Paradox, as discussed by Pearl and Mackenzie~\cite{pearl2018book}, where aggregate observational correlations mask and even invert the true conditional causal effects.

The success rates conditioned on $O$ (Figure~\ref{fig:simpson}a) show where the na\"ive estimate goes wrong. With a clear corridor the task succeeds almost regardless of the action (51/52 against 19/19) and the robot acts less often; with obstacles the context is harder and the robot acts far more often, yet the action still raises the success rate from 0.10 to 0.26 (1/10 against 5/19). Obstacles are present in $16\%$ of the episodes without a signal but in $50\%$ of those with one (Figure~\ref{fig:simpson}b), so pooling the two regimes makes the action look associated with failure, when what drives both the higher action rate and the lower success rate is the difficulty of the context. The gain stops there because clearing the crowd cannot give back the space a suitcase occupies.

\subsection{Robustness of the Estimate}

The conclusion does not hinge on the exact graph. Adding an edge $P_i \to P_e$, on the intuition that the two occupancy readings are the same quantity at two instants, makes $\{P_i, O\}$ the adjustment set; adjusting for it over the cells where both actions are observed then gives $+0.11$, again positive. Constrained hill-climbing on the 100 episodes, with $O$ and $P_i$ exogenous and $T$ a leaf, recovers $P_i \to A$, $O \to A$, $A \to P_e$, $P_e \to S$ and $O \to S$, together with that same $P_i \to P_e$, so the assumed structure is not arbitrary. The one case the data cannot speak to is a clear corridor in which the signal is nevertheless emitted, which the policy never produces.

\section{Conclusion}
This work underscores the importance of causal reasoning for social robot navigation and decision-making. While na\"{\i}ve observational conditioning suggests that emitting a social signal reduces transit success by 21 percentage points, causally adjusting for static obstacle confounding reverses the effect to a positive gain. This sign reversal illustrates how unadjusted decision-making in human-robot interaction can mislead a robot into suppressing actions that are genuinely beneficial. %

The main limitation is sample size: at 100 episodes some combinations of the variables are observed only a few times, and the one where the corridor is clear and the signal fires is empty, since the policy never acts without perceived congestion; a small exploration probability would populate it. The study is also entirely in simulation, with deterministic pedestrian compliance. 
Future work will address real-world experiments, following the human-robot spatial interaction studies in \cite{castri2023, castri2026causality, castri2024roman}, capturing probabilistic compliance and continuous spatial variations, closing the loop between causal estimation and online decision-making.

\section*{Declaration on Generative AI}
  
During the preparation of this work, the author(s) used Claude in order to: Grammar and
spelling check, Paraphrase and reword. After using this tool, the author(s) reviewed and
edited the content as needed and take(s) full responsibility for the publication's content.
\bibliography{sample-ceur}

\end{document}